\documentclass[11pt]{article}
\usepackage[a4paper,
left=1in,
right=1in,
top=1in,
bottom=1in]{geometry}
\usepackage{url}
\usepackage{orcidlink}
\usepackage{booktabs}
\usepackage{amssymb}
\usepackage{hyperref}

\usepackage{graphicx} 
\usepackage{subcaption}
\usepackage{enumitem}
\usepackage{amsmath}

\begin{document}
\title{GourNet: A CNN-Based Model for Mango Leaf Disease Detection}

\author{
Ekram Alam$^{1}$\thanks{Corresponding author: \texttt{ealam@ieee.org}}, 
Jaydip Sanyal$^{2}$, Akhil Kumar Das$^{1}$, \\
Arijit Bhattacharya$^{1}$, Farhana Sultana$^{2}$ \\[4pt]
$^{1}$Gour Mahavidyalaya, Malda, India \\
$^{2}$University of Gour Banga, Malda, India
}

\date{}


%
%

\maketitle              
\footnotetext{
This is the author’s version of the paper accepted at AdComSys 2025.
The final version is available at Springer:
\url{https://doi.org/10.1007/978-3-032-20253-6}
}
\begin{abstract}
Mango cultivation is crucial in the agricultural sector, significantly contributing to economic development and food security. However, diseases affecting mango leaves can significantly reduce both the production and overall fruit grade. Detecting leaf diseases at an early stage with precision is key to effective disease prevention and sustaining crop productivity. In this paper, we introduce a ``deep learning'' model named ``GourNet'', which leverages ``Convolutional Neural Networks'' to identify infections in mango leaves. We utilize the ``MangoLeafBD'' (MBD) dataset to train and assess the effectiveness of the presented model. The MBD dataset contains seven disease classes and a Healthy class, making a total of eight classes. To enhance model performance, the images are preprocessed through steps like resizing, rescaling, and data augmentation prior to training. To properly evaluate the model, the dataset is separated into 80\% for training, with the remaining 20\% equally split between validation and testing. Our model uses only 683,656 total parameters and achieves a classification accuracy of 97\%.
 This research’s source code can be found at: \href{https://github.com/ekramalam/GourNet-Repo}{https://github.com/ekramalam/GourNet-Repo}.


\end{abstract}
\section{Introduction}
\label{S-Introduction}
Mango, scientifically known as Mangifera indica, ranks among the most widely cultivated and valued fruit crops globally, renowned for its delicious taste and nutritional value. Worldwide production is predominantly concentrated in Asia, which accounts for 75\% of the total output, followed by South and North America with approximately a 10\% share \cite{nhbMANGO}. Extensively cultivated in tropical and subtropical regions, mango contributes notably to the agricultural sectors and provides a crucial livelihood for millions of farmers \cite{chay2019review}. Ensuring sustainable mango production is challenging due to the widespread occurrence of various leaf diseases, which significantly impact tree health and yield.

Leaf diseases in mango trees, caused by fungi, bacteria, viruses, and other pathogens, present a major risk to crop output, fruit standards, and orchard sustainability. These diseases manifest through symptoms such as leaf spots, lesions, discoloration, and defoliation, resulting in reduced photosynthetic efficiency, stunted growth, and even tree mortality \cite{arivazhagan2018mango}. Timely and accurate detection \cite{sardogan2018plant} of such diseases is essential in order to carry out successful disease prevention management strategies, preventing yield losses, and ensuring the long-term sustainability of mango cultivation.

Conventional disease detection methods typically rely on visual assessments by experienced agronomists, which can be inefficient, heavily reliant on manual work, and affected by personal bias. With advancements in ``Computer Vision'' (CV) \cite{alam2021leveraging} and ``Deep Learning'' (DL) based approaches \cite{bengio2017deep}, DL-based automated systems \cite{gulavnai2019deep} have shown great potential in identifying plant diseases. These systems use publicly available datasets of diseased and healthy leaves to train models that can accurately classify leaves as either diseased or healthy.

This study makes use of the ``MangoLeafBD" (MBD) dataset, introduced by Ahmed et al. \cite{ahmed2023mangoleafbd}. ``GourNet'', a Convolutional Neural Network (CNN) -based model presented in this paper, serves as an effective tool for in-time detection and proactive handling of mango leaf diseases (MLDs) in orchards. This empowers farmers and agricultural stakeholders to take informed actions, thereby minimizing disease impact on crop productivity and ensuring sustainable cultivation practices.

This paper is structured in the following manner: Section \ref{S_Background_study} reviews the foundational concepts relevant to this work. Section \ref{S_Related_work} discusses prior research conducted in this area. Section \ref{S_Dataset} introduces the MBD dataset employed in the experiments. Section \ref{S_Methodology} elaborates on the proposed method for detecting diseases in mango leaves. Section \ref{S_Results} describes the experimental framework and analyzes the findings. Finally, Section \ref{S_conclusion} wraps up the study and discusses possible future enhancements.

\section{Background Study}
\label{S_Background_study}
Recent years have seen remarkable progress in Artificial Intelligence (AI) \cite{varghese2024artificial}, especially driven by developments in DL \cite{chen2025deep}. Among the diverse DL techniques, CNNs \cite{alzubaidi2021review}\cite{lei2019dilated} have demonstrated remarkable effectiveness for image classification, showing outstanding capabilities in pattern recognition. CNNs leverage backpropagation and optimization techniques to iteratively refine their parameters, minimizing the LF by reducing discrepancies between predicted and actual labels. 
One notable strength of CNNs lies in their capability to automatically extract hierarchical features—initial layers identify basic elements like edges and textures, whereas deeper layers recognize higher-level, more abstract patterns.

Owing to their effectiveness in handling spatial data, CNNs  have revolutionized CV applications, attaining top-tier accuracy in multiple areas, including ``image classification'', ``facial recognition'', ``object detection'', and ``medical image analysis'' \cite{alam2021leveraging,upadhyay2025deep}. Their robust feature extraction capabilities make them indispensable for both research and real-world applications.

This work introduces ``GourNet'', a CNN-based model designed for mango leaf disease detection (MLDD). By leveraging various layers of the network, GourNet automatically extracts discriminative features from leaf images, enabling precise classification of different disease types. Figure \ref{fig:cnn} shows a typical CNN model. The different layers of a CNN model are briefly discussed below.
\begin{itemize}[label=\textbullet]
    
\begin{figure}[htbp]
    \centering
    \includegraphics[width=0.8\textwidth]{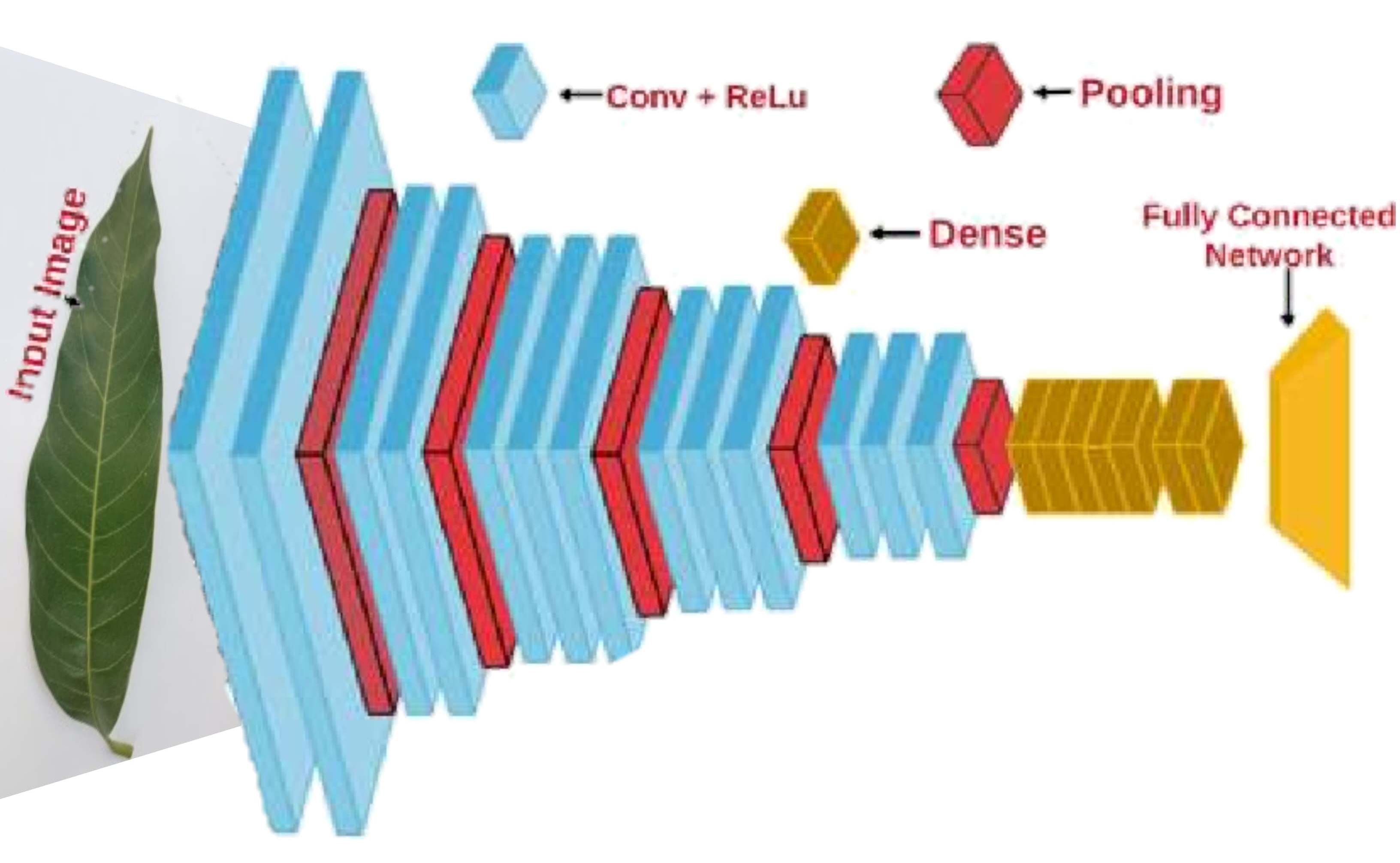} 
    \caption{Structural Overview of a typical CNN Model}
    \label{fig:cnn}
\end{figure}

\item  \textbf{Convolutional Layers (CLs):} CLs use trainable kernels to scan input images and capture spatial features. These filters facilitate the extraction of local visual cues, including edges, textures, and shapes. As the data progresses through deeper layers, the features become increasingly abstract and representative of higher-level structures \cite{bhatt2021cnn}.
 
 \item  \textbf{Activation Function (AF):}
To introduce non-linearity into the model, AFs like ``Rectified Linear Unit'' (ReLU) \cite{ramadevi2024fractional} are commonly used following convolutional operations.
This non-linear transformation allows the network to capture complex and non-trivial characteristic features of the dataset, which linear models cannot capture effectively.

 \item \textbf{Pooling Layer (PL):}
PLs, such as MaxPooling2D, are employed to reduce the spatial size of feature maps. These layers, often employing a $2 \times 2$ window, downsample the feature maps to reduce computational complexity and enhance the model's resilience to minor shifts or deformations in the input \cite{chua1993cnn}.

\item \textbf{Fully Connected (FC) Layers:}
Once spatial features have been obtained, they are reshaped into a 1D vector and passed into FC layers. These layers perform higher-level interpretation and make predictions using the learned features.














\item \textbf{Softmax Function:} The softmax function is typically applied in the final classification layer to transform output scores into probability values. This enables the model to assign a likelihood to each class and make a final prediction based on the highest probability.

Given an input vector $\mathbf{z}$ containing $K$ elements, the softmax function converts each element $z_i$ into a probability asdefined in Equation \ref{eq:softmax} \cite{sufian2021deep}.

\begin{equation}
\text{softmax}(z_i) = \frac{e^{z_i}}{\sum_{j=1}^{K} e^{z_j}}
\label{eq:softmax}
\end{equation}

where:
\begin{itemize}
    \item $i$ is the index of the current element (from $1$ to $K$),
    \item $j$ iterates over all elements in the vector,
    \item $e$ denotes the exponential function.
\end{itemize}

Suppose a CNN processes an image and produces an output vector $\mathbf{z}$ with 8 elements, each representing a class score. Suppose the value of the vector $\mathbf{z}$ is as provided in Equation \ref{eq:z_vector}.

\begin{equation}
\mathbf{z} = [2.5, -1.2, 4.1, 0.8, -0.3, 3.7, 1.9, -2.0]
\label{eq:z_vector}
\end{equation}

Next, the exponentials of the elements in the vector $\mathbf{z}$ are computed. The resulting values are presented in Equation~\ref{eq:exp}.

\begin{equation}
\exp(\mathbf{z}) = [12.182, 0.301, 60.340, 2.226, 0.741, 40.447, 6.686, 0.135]
\label{eq:exp}
\end{equation}

Next, the sum of these exponentials is calculated as mentioned in Equation \ref{eq:Sum_exp}.

\begin{equation}
\sum \exp(\mathbf{z}) = 122.958
\label{eq:Sum_exp}
\end{equation}

Finally, each exponentiated value is divided by the total sum to yield the softmax output as shown in Equation \ref{eq:SoftOp}.

\begin{equation}
\text{softmax}(\mathbf{z}) = [0.098, 0.002, 0.491, 0.018, 0.006, 0.329, 0.054, 0.001]
\label{eq:SoftOp}
\end{equation}

The output vector indicates the likelihood for each of the eight classes, and the class with the highest probability (0.491) is selected as the model’s final prediction.

\end{itemize}

\section{Previous Studies}
\label{S_Related_work}
This section offers a Summary of research efforts that utilize machine learning (ML), with a focus on DL, for detecting diseases in mango leaves.
  
 Rizvee et al.~\cite{rizvee2023leafnet} proposed a CNN-based model, LeafNet, for detecting MLDs using the MBD dataset. While their model contains 3,256,608 total parameters, the proposed GourNet architecture requires only 683,656. Puranik et al. \cite{puranik2024mobilenetv3} employed the MobileNetV3 architecture for MLDD using the MBD dataset. Pandiyaraju et al.~\cite{pandiyaraju2024mango} proposed an ensemble learning approach that combines VGG-16 and EfficientNetV2-B0 with a spatial attention mechanism for classifying MLDs. Pradhan et al \cite{pradhan2024deep} employed the EfficientNet model to detect and classify various MLDs. Gowrishankar et al. \cite{gowrishankar2023convnext} leveraged ConvNeXt models to identify diseases in mango plants caused by pathogens and pests. 
 
 Mahbub et al. \cite{mahbub2023detect} proposed a lightweight CNN for accurately classifying seven different MLDs along with normal mango leaves. Soni et al. \cite{sonimango} propose two deep learning architectures, Shallow CNN and Residual CNN, for detecting seven mango diseases and healthy plants using leaf images. Chandrasekar et al. \cite{chandrashekarmdcn} introduced a DenseNet architecture model with DL to detect MLDs using the MBD dataset. Mahmud et al. \cite{mahmud2024light} introduced a lightweight, DL-based technique for classifying MLDs using a customized DenseNet architecture.

 Hasana et al. \cite{hasana2023speeding} explored the application of genetic algorithms with ``EfficientNetB0'' pre-trained on ``ImageNet'' and utilizing ``Food-101'', ``CIFAR-100'', and MBD datasets \cite{alam2024gmdcsa}. Rahman et al. \cite{rahman2023novel} employed a ML model for disease identification in mango cultivation, using a Gradient Boosting Classifier with image segmentation and feature extraction techniques. Giri et al. \cite{giri2023enhancing} applied a Gradient Boosting Classifier to classify two MLDs— ``Powdery Mildew'' \cite{alkolaly2022impact} and ``Sooty Mould'' \cite{alkolaly2022impact} —using the MBD dataset. The method utilized ``mean shift segmentation'' along with ``Hu moments'' to extract features from the images.

Ramadan et al. \cite{ramadan2023enhancing} employ CycleGAN-generated data augmentation to classify MLDs. Garg et al. \cite{garg2023systematic} researched to improve the identification and diagnosis of plant diseases, particularly on mango leaves, by utilizing image processing techniques. This involves preprocessing images, segmenting leaf components, extracting features, and classifying diseases.

Rayed et al.\cite{rayed2023vision}, Salamai et al.\cite{salamai2023enhancing}, and Hossain et al. \cite{hossain2024deep} investigated the use of ``Vision Transformer'' \cite{han2022survey} (ViT)-based models for MLDD using the MBD dataset. Alamri et al. \cite{alamri2025mango} explored the use of ConvNeXt and ViT architectures for diagnosing mango diseases using two datasets: MBD for leaf diseases and SenMangoFruitDDS for fruit diseases. 

\section{Dataset}
\label{S_Dataset}
\begin{figure}[h]
    \centering
    \captionsetup[subfigure]{labelformat=simple} 
    
    \begin{subfigure}{0.22\textwidth}
        \centering
        \includegraphics[width=\linewidth]{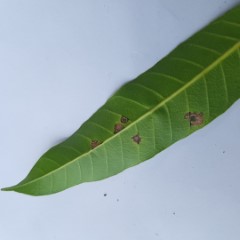}
      \caption{Representative sample image from ``Anthracnose'' class}
        \label{fig:subfig1}
    \end{subfigure}
    \hfill
    \begin{subfigure}{0.22\textwidth}
        \centering
        \includegraphics[width=\linewidth]{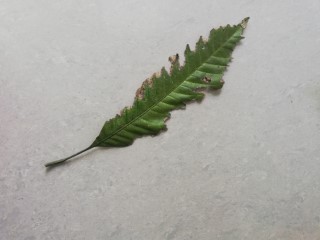}
        \caption{Representative sample image from ``Bacterial Canker'' class}
       
        \label{fig:subfig2}
    \end{subfigure}
    \hfill
    \begin{subfigure}{0.22\textwidth}
        \centering
        \includegraphics[width=\linewidth]{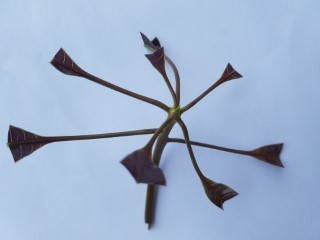}
        \caption{Representative sample image from ``Cutting Weevil'' class}
       
        \label{fig:subfig3}
    \end{subfigure}
    \hfill
    \begin{subfigure}{0.22\textwidth}
        \centering
        \includegraphics[width=\linewidth]{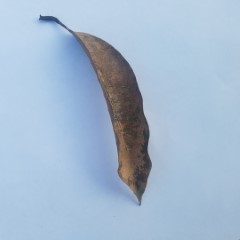}
        \caption{Representative sample image from ``Die Back'' class}
       
        \label{fig:subfig4}
    \end{subfigure}

    \vspace{0.5cm} 
    \begin{subfigure}{0.22\textwidth}
        \centering
        \includegraphics[width=\linewidth]{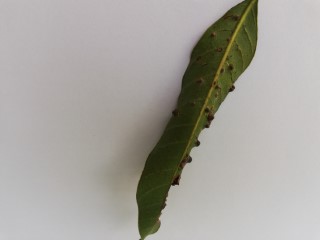}
        \caption{Representative sample image from ``Gall Midge'' class}
       
        \label{fig:subfig5}
    \end{subfigure}
    \hfill
    \begin{subfigure}{0.22\textwidth}
        \centering
        \includegraphics[width=\linewidth]{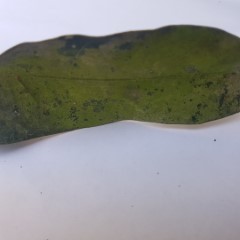}
         \caption{Representative sample image from ``Powdery Mildew'' class}
       
        \label{fig:subfig6}
    \end{subfigure}
    \hfill
    \begin{subfigure}{0.22\textwidth}
        \centering
        \includegraphics[width=\linewidth]{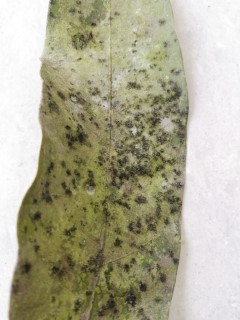}
        \caption{Representative sample image from ``Sooty Mould'' class}
        
        \label{fig:subfig7}
    \end{subfigure}
    \hfill
    \begin{subfigure}{0.22\textwidth}
        \centering
        \includegraphics[width=\linewidth]{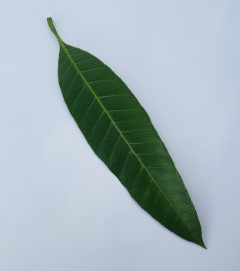}
         \caption{Representative sample image from Healthy class}
      
        \label{fig:subfig8}
    \end{subfigure}

    \caption{Sample images from the eight classes of the MBD dataset}
    \label{F:classes}
\end{figure}

The MBD dataset comprises of 4,000 mango leaf images captured at varying resolutions. In this experiment, the MBD is partitioned into training, validation, and testing sets using an 80\%, 10\%, and 10\% split, respectively.
Figure \ref{F:classes} displays representative sample images from all eight categories included in the dataset. 
 A summarized overview of these eight classes is presented below.

\begin{enumerate}
    \item \textbf{Anthracnose:} This fungal infection, triggered by ``Colletotrichum gloeosporioides'', results in the formation of dark, sunken spots on mango leaves. Other parts of the tree may also be affected, such as fruits and flowers.
   \item \textbf{Bacterial Canker:} This disease affects various parts of the mango plant, including leaves, petioles, stems, twigs, and fruits. It begins with water-soaked spots that eventually develop into characteristic lesions.
    \item \textbf{Cutting Weevil:} The most obvious signs of infestation are cut leaves on the ground under trees and stripped leafless buds that can be seen from a distance.
    
   \item \textbf{Dieback:} This condition involves the progressive drying and death of twigs and branches. It may result from fungal pathogens, lack of essential nutrients, or adverse environmental conditions.
   
\item \textbf{Gall Midge:} The larvae of gall midges consume plant tissues, leading to the formation of abnormal swellings known as galls, which can harm mango leaves, blossoms, fruits, and shoots.
    
 \item \textbf{Sooty Mould:} This fungal infection results in a dark, soot-like layer forming on mango tree leaves and stems, typically developing due to insect secretions or infestations.

    \item \textbf{Powdery Mildew:} The fungus ``Oidium mangifferae'' causes powdery mildew. Consequently, a white, powder-like layer develops on the surface of mango leaves, reducing their efficiency in carrying out photosynthesis.
    
    \item \textbf{Healthy:} This category comprises images of mango leaves that show no apparent symptoms of infection or physical damage. The leaves appear green, intact and exhibit natural texture and coloration, representing healthy plant conditions.

\end{enumerate}

Each class contains 500 JPEG (RGB) images, except for the Bacterial Canker class, which contains 501 images. The dataset comprises seven types of leaf diseases along with a category for healthy leaves.

\section{Methodology}
\label{S_Methodology}
\begin{figure}[h!]
    \centering
    \includegraphics[width=.9\textwidth]{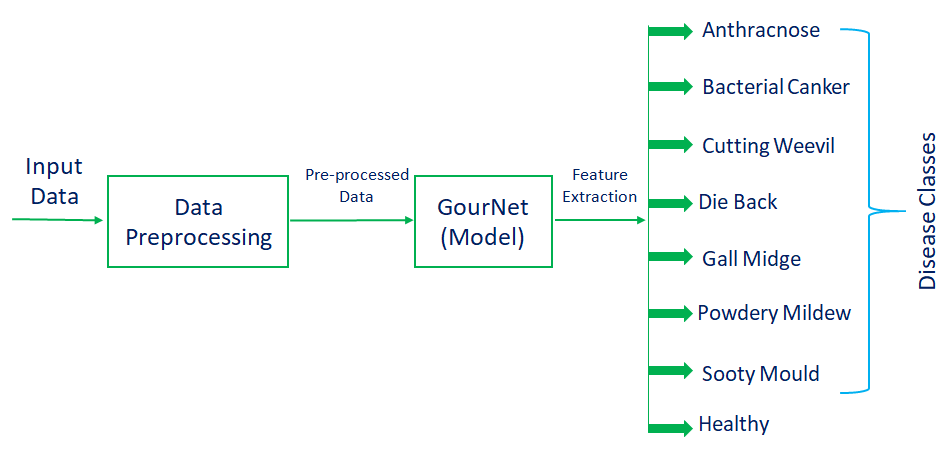} 
    \caption{Workflow of the proposed approach}
    \label{fig:method}
\end{figure}
This section presents the structured methodology adopted for classifying MLDs. Our approach includes data collection, preprocessing, model construction, and disease classification, as depicted in Figure \ref{fig:method}.

\begin{enumerate}

    \item \textbf{Data Preprocessing:} The MBD dataset \cite{ahmed2023mangoleafbd} is used in this work. The images were taken using a smartphone camera across four distinct mango orchards located in Bangladesh. This phase involves resizing, normalization, and data augmentation. Initially, each image is scaled to a consistent resolution of 224 $\times$ 224. Subsequently, the pixel intensities are scaled down from the original range of 0–255 to a normalized scale of 0–1. Finally, random flipping and rotation are employed as augmentation strategies to enhance the variety within the input data, which in turn aids the model in achieving better generalization on unfamiliar images.

    \item \textbf{Model:} The processed images are subsequently supplied to the ``GourNet'' architecture for deriving relevant features. The layer-wise architecture of the “GourNet” model is presented in Figure \ref{fig:Gnet}. In this experiment, 2D convolution and max pooling have been used.
    \begin{figure}[h!]
    \centering
    \includegraphics[width=1.05\textwidth]{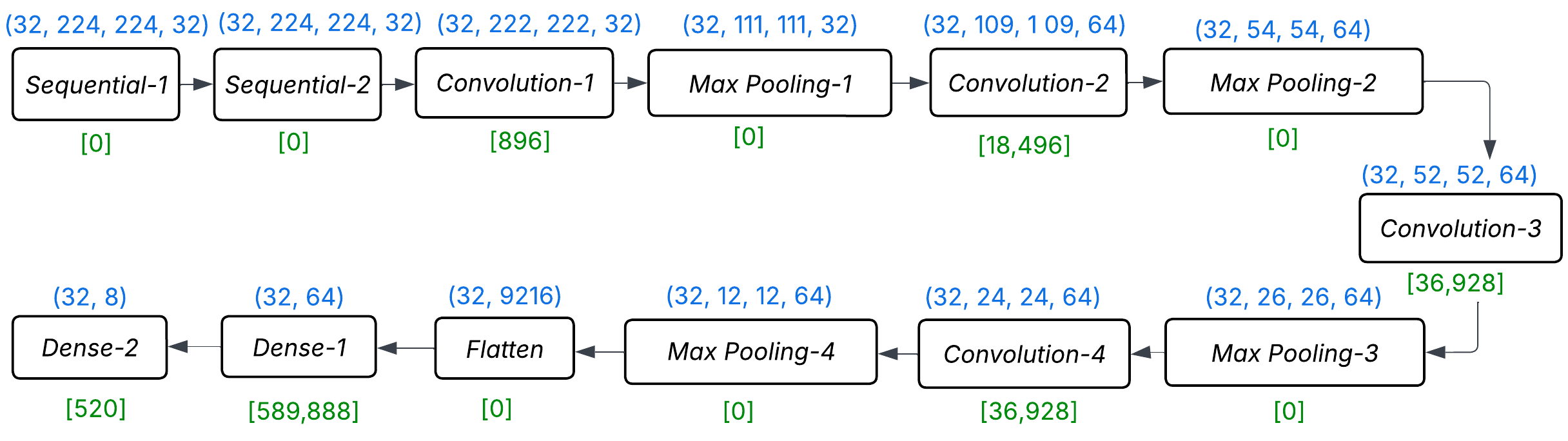} 
   \caption{Layer-wise structure of the GourNet model with output shapes and number of parameters}

    \label{fig:Gnet}
\end{figure}
Each block represents a layer; at the top of each block, the output size is shown, and at the bottom, the number of parameters is indicated.
    Each input image is resized to $224 \times 224$ pixels with three color channels. 
    
    GourNet follows a sequential architecture starting with a 2D CL consisting of 32 filters of dimension $3 \times 3$ and employs the ReLU AF. 
    
    Next, a $2 \times 2$ max pooling operation is performed. The subsequent convolutional and PLs deepen the network and capture higher-level features. The resulting feature maps are reshaped into a 1D vector and input into a FC layer with ReLU AF. The final classification layer contains 8 neurons and applies the softmax function to handle multiclass output. The Adam optimizer (AO) was used in this model. 

    \item \textbf{Disease Classification:} Finally, the GourNet model classifies the input image into one of the eight classes using the softmax classifier.
\end{enumerate}

\section{Experimental Setup and Results}
   \label{S_Results}
\begin{table}[h!]
\centering
\caption{Versions of Python interpreter and libraries used in this work}
\label{tab:packages}
\begin{tabular}{p{3cm}p{2.5cm}p{3cm}p{2.5cm}}
\hline
\textbf{Package Name} & \textbf{Version} & \textbf{Package Name} & \textbf{Version} \\
\hline
Python      & 3.11.4 & NumPy       & 1.26.4 \\
Keras       & 3.2.1  & TensorFlow  & 2.16.1 \\
Matplotlib  & 3.8.3  &             &        \\
\hline
\end{tabular}
\end{table}
This experiment was performed using a laptop featuring an Intel Core i3-1115G4 (11th Gen) CPU of frequency 3.00 GHz, 8 GB of RAM, and Intel UHD Graphics (128 MB). Table~\ref{tab:packages} lists the Python interpreter and library versions used in this work.


Table~\ref{tab:training_config} displays the training configurations used for the proposed \texttt{GourNet} model, including the learning rate (LR), loss function (LF), optimizer, and other details.
\begin{table}[htbp]
\centering
\caption{Model Training Configuration}
\label{tab:training_config}
\begin{tabular}{p{3cm}p{7cm}}
\hline
\textbf{Parameter} & \textbf{Value} \\
\hline
LR & 0.001 \\
LF & ``Sparse Categorical Cross-Entropy'' \\
Optimizer & AO \\

Early Stopping & Enabled (monitored on \texttt{val\_loss} \\
Patience & 3 \\
Batch Size & 32 \\
Maximum Epochs & 50 \\
\hline
\end{tabular}
\end{table}

\begin{figure}[htbp]
    \centering
    \includegraphics[width=1\textwidth]{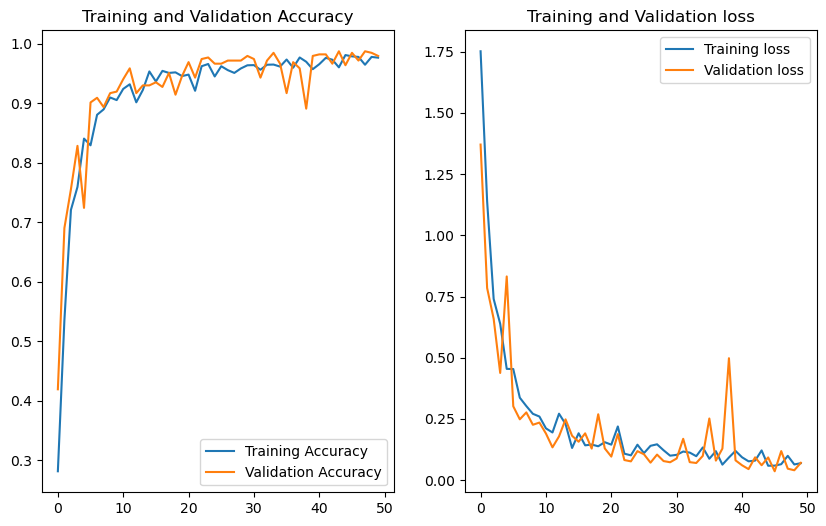} 
    \caption{Training and validation accuracy and loss trends of the GourNet model}
    \label{fig:result}
\end{figure}

Figure \ref{fig:result} illustrates the accuracy and loss trends across epochs for both training and validation data. The GourNet model was trained only for 50 epochs. Both training and validation accuracy exhibit a steady upward trend across epochs, highlighting the model’s strong learning ability and generalization performance. Similarly, both training and validation loss curves exhibit a steady decline, suggesting that GourNet successfully minimized the LF without signs of overfitting. GourNet achieved a classification accuracy of 97\% using only 683,656 parameters.
Table~\ref{tab:params} presents a comparative analysis of total and trainable parameters among various models, including the proposed GourNet.

\begin{table}[h!]
\centering
\caption{Comparison of total and trainable parameters across architectures}
\label{tab:params}
\begin{tabular}{p{4cm}p{3cm}p{4cm}}
\toprule
\textbf{Architecture Name} & \textbf{Total Parameters} & \textbf{Trainable Parameters} \\
\midrule
GourNet   & 683,656    & 683,656 \\
LeafNet   & 3,256,608  & 3,255,628 \\
AlexNet   & 58,319,624 & 58,316,872 \\
VGG16     & 134,293,320 & 134,293,320 \\
\bottomrule
\end{tabular}
\end{table}

\section{Conclusion and Future Scope}
  \label{S_conclusion}
The proposed GourNet model demonstrates good performance in classifying various MLDs and healthy samples using a reduced number of parameters. Unlike traditional machine learning algorithms, GourNet avoids manual feature engineering and ensures rapid convergence with high accuracy. The proposed model can be generalized and fine-tuned to classify leaf diseases in other plants and trees. 

%







\bibliographystyle{plain} 
 \bibliography{mldd}

@article{rizvee2023leafnet,
  title={LeafNet: A proficient convolutional neural network for detecting seven prominent mango leaf diseases},
  author={Rizvee, Redwan Ahmed and Orpa, Tasnim Hossain and Ahnaf, Adil and Kabir, Md Ahsan and Rashid, Mohammad Rifat Ahmmad and Islam, Mohammad Manzurul and Islam, Maheen and Jabid, Taskeed and Ali, Md Sawkat},
  journal={Journal of Agriculture and Food Research},
  volume={14},
  pages={100787},
  year={2023},
  publisher={Elsevier}
}

@article{sufian2021deep,
  title={Deep learning in computer vision through mobile edge computing for iot},
  author={Sufian, Abu and Alam, Ekram and Ghosh, Anirudha and Sultana, Farhana and De, Debashis and Dong, Mianxiong},
  journal={Mobile Edge Computing},
  pages={443--471},
  year={2021},
  publisher={Springer}
}

@misc{nhbMANGO,
	author = {},
	title = {{M}{A}{N}{G}{O} --- nhb.gov.in},
	howpublished = {\url{https://nhb.gov.in/report_files/mango/mango.htm}},
	year = {},
	note = {[Accessed 15-06-2025]},
}

@article{chen2025deep,
  title={Deep learning-based software engineering: progress, challenges, and opportunities},
  author={Chen, Xiangping and Hu, Xing and Huang, Yuan and Jiang, He and Ji, Weixing and Jiang, Yanjie and Jiang, Yanyan and Liu, Bo and Liu, Hui and Li, Xiaochen and others},
  journal={Science China Information Sciences},
  volume={68},
  number={1},
  pages={1--88},
  year={2025},
  publisher={Springer}
}

@article{ramadevi2024fractional,
  title={Fractional ordering of activation functions for neural networks: A case study on Texas wind turbine},
  author={Ramadevi, Bhukya and Kasi, Venkata Ramana and Bingi, Kishore},
  journal={Engineering Applications of Artificial Intelligence},
  volume={127},
  pages={107308},
  year={2024},
  publisher={Elsevier}
}

@article{varghese2024artificial,
  title={Artificial intelligence in surgery},
  author={Varghese, Chris and Harrison, Ewen M and O’Grady, Greg and Topol, Eric J},
  journal={Nature medicine},
  volume={30},
  number={5},
  pages={1257--1268},
  year={2024},
  publisher={Nature Publishing Group US New York}
}

@book{bengio2017deep,
  title={Deep learning},
  author={Bengio, Yoshua and Goodfellow, Ian and Courville, Aaron and others},
  volume={1},
  year={2017},
  publisher={MIT press Cambridge, MA, USA}
}

@article{arivazhagan2018mango,
  title={Mango leaf diseases identification using convolutional neural network},
  author={Arivazhagan, S and Ligi, S Vineth},
  journal={International Journal of Pure and Applied Mathematics},
  volume={120},
  number={6},
  pages={11067--11079},
  year={2018}
}

@article{ahmed2023mangoleafbd,
  title={MangoLeafBD: A comprehensive image dataset to classify diseased and healthy mango leaves},
  author={Ahmed, Sarder Iftekhar and Ibrahim, Muhammad and Nadim, Md and Rahman, Md Mizanur and Shejunti, Maria Mehjabin and Jabid, Taskeed and Ali, Md Sawkat},
  journal={Data in Brief},
  volume={47},
  pages={108941},
  year={2023},
  publisher={Elsevier}
}

@article{gowrishankar2023convnext,
  title={Convnext-based mango leaf disease detection: Differentiating pathogens and pests for improved accuracy},
  author={Gowrishankar, S},
  journal={International Journal of Advanced Computer Science and Applications},
  volume={14},
  number={6},
  year={2023},
  publisher={Science and Information (SAI) Organization Limited}
}

@article{hasana2023speeding,
  title={Speeding Up EfficientNet: Selecting Update Blocks of Convolutional Neural Networks using Genetic Algorithm in Transfer Learning},
  author={Hasana, Md Mehedi and Ibrahim, Muhammad and Ali, Md Sawkat},
  journal={arXiv preprint arXiv:2303.00261},
  year={2023}
}

@article{salamai2023enhancing,
  title={Enhancing mango disease diagnosis through eco-informatics: A deep learning approach},
  author={Salamai, Abdullah Ali},
  journal={Ecological Informatics},
  volume={77},
  pages={102216},
  year={2023},
  publisher={Elsevier}
}

@article{rahman2023novel,
  title={A Novel Neural Network-Based Federated Learning System for Imbalanced and Non-IID Data},
  author={Rahman Chowdhury, Mahfuzur and Ibrahim, Muhammad},
  journal={arXiv e-prints},
  pages={arXiv--2311},
  year={2023}
}

@inproceedings{rayed2023vision,
  title={A vision transformer-based approach for recognizing seven prevalent mango leaf diseases},
  author={Rayed, Md Eshmam and Alfaz, Nazia and Niha, Sadia Islam and Islam, SM Sajibul and others},
  booktitle={2023 26th International Conference on Computer and Information Technology (ICCIT)},
  pages={1--6},
  year={2023},
  organization={IEEE}
}

@inproceedings{mahbub2023detect,
  title={Detect bangladeshi mango leaf diseases using lightweight convolutional neural network},
  author={Mahbub, Nosin Ibna and Naznin, Feroza and Hasan, Md Imran and Shifat, Syed Mahfuzur Rahman and Hossain, Md Alamgir and Islam, Md Zahidul},
  booktitle={2023 International Conference on Electrical, Computer and Communication Engineering (ECCE)},
  pages={1--6},
  year={2023},
  organization={IEEE}
}

@inproceedings{ramadan2023enhancing,
  title={Enhancing Mango Leaf Disease Classification: ViT, BiT, and CNN-Based Models Evaluated on CycleGAN-Augmented Data},
  author={Ramadan, Syed Taha Yeasin and Sakib, Tanjim and Rahat, Md Ahsan and Mosharrof, Shakil and Rakin, Fatin Ishrak and Jahangir, Raiyan},
  booktitle={2023 26th International Conference on Computer and Information Technology (ICCIT)},
  pages={1--6},
  year={2023},
  organization={IEEE}
}

@inproceedings{garg2023systematic,
  title={A systematic analysis of various techniques for mango leaf disease detection},
  author={Garg, Rinku and Sandhu, Amanpreet Kaur and Kaur, Bobbinpreet},
  booktitle={2023 International Conference on Disruptive Technologies (ICDT)},
  pages={349--354},
  year={2023},
  organization={IEEE}
}

@article{sonimango,
  title={MANGO DISEASE CLASSIFICATION-SHALLOW CNN OR RESIDUAL CNN?-WHICH IS BETTER?},
  author={Soni, Dhruv and Jumadinoda, Janyl}
}

@article{chandrashekarmdcn,
  title={MDCN: Modified Dense Convolution Network Based Disease Classification in Mango Leaves},
  author={Chandrashekar, Chirag and Vijayakumar, KP and Pradeep, K and Balasundaram, A}
}

@article{mahmud2024light,
  title={Light-Weight Deep Learning Model for Accelerating the Classification of Mango-Leaf Disease},
  author={Mahmud, Bahar Uddin and Al Mamun, Abdullah and Hossen, Md Jakir and Hong, Guan Yue and Jahan, Busrat},
  journal={Emerging Science Journal},
  volume={8},
  number={1},
  pages={28--42},
  year={2024}
}

@article{gulavnai2019deep,
  title={Deep learning for image based mango leaf disease detection},
  author={Gulavnai, Sampada and Patil, Rajashri},
  journal={International Journal of Recent Technology and Engineering},
  volume={8},
  number={3S3},
  pages={54--56},
  year={2019}
}

@inproceedings{sardogan2018plant,
  title={Plant leaf disease detection and classification based on CNN with LVQ algorithm},
  author={Sardogan, Melike and Tuncer, Adem and Ozen, Yunus},
  booktitle={2018 3rd international conference on computer science and engineering (UBMK)},
  pages={382--385},
  year={2018},
  organization={IEEE}
}

@article{alzubaidi2021review,
  title={Review of deep learning: concepts, CNN architectures, challenges, applications, future directions},
  author={Alzubaidi, Laith and Zhang, Jinglan and Humaidi, Amjad J and Al-Dujaili, Ayad and Duan, Ye and Al-Shamma, Omran and Santamar{\'\i}a, Jos{\'e} and Fadhel, Mohammed A and Al-Amidie, Muthana and Farhan, Laith},
  journal={Journal of big Data},
  volume={8},
  pages={1--74},
  year={2021},
  publisher={Springer}
}

@article{lei2019dilated,
  title={A dilated CNN model for image classification},
  author={Lei, Xinyu and Pan, Hongguang and Huang, Xiangdong},
  journal={IEEE Access},
  volume={7},
  pages={124087--124095},
  year={2019},
  publisher={IEEE}
}

@article{bhatt2021cnn,
  title={CNN variants for computer vision: History, architecture, application, challenges and future scope},
  author={Bhatt, Dulari and Patel, Chirag and Talsania, Hardik and Patel, Jigar and Vaghela, Rasmika and Pandya, Sharnil and Modi, Kirit and Ghayvat, Hemant},
  journal={Electronics},
  volume={10},
  number={20},
  pages={2470},
  year={2021},
  publisher={MDPI}
}

@article{chua1993cnn,
  title={The CNN paradigm},
  author={Chua, Leon O and Roska, Tamas},
  journal={IEEE Transactions on Circuits and Systems I: Fundamental Theory and Applications},
  volume={40},
  number={3},
  pages={147--156},
  year={1993},
  publisher={IEEE}
}

@article{chay2019review,
  title={A review on production and marketing of mango fruit},
  author={Chay, Kayier Guien and Workeneh, Amsale and Shifera, Beshadu},
  journal={World Journal of Agriculture and Soil Science},
  volume={2},
  number={2},
  pages={1--7},
  year={2019}
}

@inproceedings{alam2021leveraging,
  title={Leveraging deep learning for computer vision: A review},
  author={Alam, Ekram and Sufian, Abu and Das, Akhil Kumar and Bhattacharya, Arijit and Ali, Md Firoj and Rahman, MM Hafizur},
  booktitle={2021 22nd International Arab Conference on Information Technology (ACIT)},
  pages={1--8},
  year={2021},
  organization={IEEE}
}

@inproceedings{puranik2024mobilenetv3,
  title={MobileNetV3 for Mango Leaf Disease Detection: An efficient Deep Learning Approach for Precision Agriculture},
  author={Puranik, Sukruth S and Hanamakkanavar, Siddharth R and Bidargaddi, Anupama P and Ballur, Vighnesh V and Joshi, Pratham T and SM, Meena and Kulkarni, Uday},
  booktitle={2024 5th International Conference for Emerging Technology (INCET)},
  pages={1--7},
  year={2024},
  organization={IEEE}
}

@article{giri2023enhancing,
  title={Enhancing disease management in mango cultivation: A machine learning approach to classifying leaf diseases},
  author={Giri, Gst Ayu Vida Mastrika and Musdar, Izmy Alwiah and Angriani, Husni and Taruk, Medi},
  journal={Indonesian Journal of Data and Science},
  volume={4},
  number={3},
  pages={160--168},
  year={2023}
}

@inproceedings{pandiyaraju2024mango,
  title={Mango Leaf Disease Detection and Classification Using Spatial Attention Enabled Ensemble Classification},
  author={Pandiyaraju, V and Venkatraman, Shravan and Abeshek, A and Aravintakshan, SA and others},
  booktitle={2024 International Conference on Advances in Data Engineering and Intelligent Computing Systems (ADICS)},
  pages={1--8},
  year={2024},
  organization={IEEE}
}

@article{hossain2024deep,
  title={Deep learning for mango leaf disease identification: A vision transformer perspective},
  author={Hossain, Md Arban and Sakib, Saadman and Abdullah, Hasan Muhammad and Arman, Shifat E},
  journal={Heliyon},
  volume={10},
  number={17},
  year={2024},
  publisher={Elsevier}
}

@inproceedings{pradhan2024deep,
  title={A Deep Learning Based Approach for Detecting Mango Leaf Diseases},
  author={Pradhan, Astik Kumar and Rout, Jitendra Kumar and Ghosh, Debanjan and Sinha, Shailendra Kumar},
  booktitle={2024 6th International Conference on Computational Intelligence and Networks (CINE)},
  pages={1--6},
  year={2024},
  organization={IEEE}
}

@article{alamri2025mango,
  title={Mango Disease Detection Using Fused Vision Transformer with ConvNeXt Architecture.},
  author={Alamri, Faten S and Sadad, Tariq and Almasoud, Ahmed S and Aurangzeb, Raja Atif and Khan, Amjad},
  journal={Computers, Materials \& Continua},
  volume={83},
  number={1},
  year={2025}
}

@article{upadhyay2025deep,
  title={Deep learning and computer vision in plant disease detection: a comprehensive review of techniques, models, and trends in precision agriculture},
  author={Upadhyay, Abhishek and Chandel, Narendra Singh and Singh, Krishna Pratap and Chakraborty, Subir Kumar and Nandede, Balaji M and Kumar, Mohit and Subeesh, A and Upendar, Konga and Salem, Ali and Elbeltagi, Ahmed},
  journal={Artificial Intelligence Review},
  volume={58},
  number={3},
  pages={92},
  year={2025},
  publisher={Springer}
}

@article{alam2024gmdcsa,
  title={GMDCSA-24: a dataset for human fall detection in videos},
  author={Alam, Ekram and Sufian, Abu and Dutta, Paramartha and Leo, Marco and Hameed, Ibrahim A},
  journal={Data in Brief},
  volume={57},
  pages={110892},
  year={2024},
  publisher={Elsevier}
}

@article{alkolaly2022impact,
  title={Impact of powdery mildew and sooty mold diseases on mango by natural fungicide},
  author={Alkolaly, Asmaa M and Hassan, Rodina A and Monir, Gehan A},
  journal={International Journal of Scientific Research Updates},
  volume={3},
  number={2},
  pages={127--133},
  year={2022}
}

@article{han2022survey,
  title={A survey on vision transformer},
  author={Han, Kai and Wang, Yunhe and Chen, Hanting and Chen, Xinghao and Guo, Jianyuan and Liu, Zhenhua and Tang, Yehui and Xiao, An and Xu, Chunjing and Xu, Yixing and others},
  journal={IEEE transactions on pattern analysis and machine intelligence},
  volume={45},
  number={1},
  pages={87--110},
  year={2022},
  publisher={IEEE}
}

\end{document}